\documentclass[10pt,twocolumn,letterpaper]{article}

\usepackage{cvpr}              
\definecolor{cvprblue}{rgb}{0.21,0.49,0.74}
\usepackage[pagebackref,breaklinks,colorlinks,allcolors=cvprblue]{hyperref}

\def\paperID{} 
\def\confName{CVPR}
\def\confYear{2026}

\title{PatchScene: Patch-based Voxel Diffusion for Large-Scale Scene Completion}

\author{
Qingdong Xu$^{1,*}$ \quad
Jiajun Zhu$^{1,2,*}$ \quad
Shilin Zhu$^{4,*}$ \quad \\
Xinjing He$^{5}$ \quad
Chao Lu$^{2}$ \quad
Huanran Wang$^{2}$ \quad
Jiyao Zhang$^{3,\dagger}$ \\
\\
{\normalsize
$^{1}$MEGVII Technology \quad
$^{2}$Qianli Technology \quad
$^{3}$Peking University }\\
{\normalsize
$^{4}$Northeastern University, China \quad
$^{5}$Northwest Polytechnical University, Xi'an } \\
{\tt\small \{2301010,2301062\}@stu.neu.edu.cn \quad mr.zhujiajun@gmail.com \quad jiyaozhang@stu.pku.edu.cn} \\
\small{$^*$Equal contribution \quad $^\dagger$Corresponding author}
}

\begin{document}
\raggedbottom
\twocolumn[{
\maketitle
\vspace{-3.5em}
\begin{center}
\includegraphics[width=\textwidth]{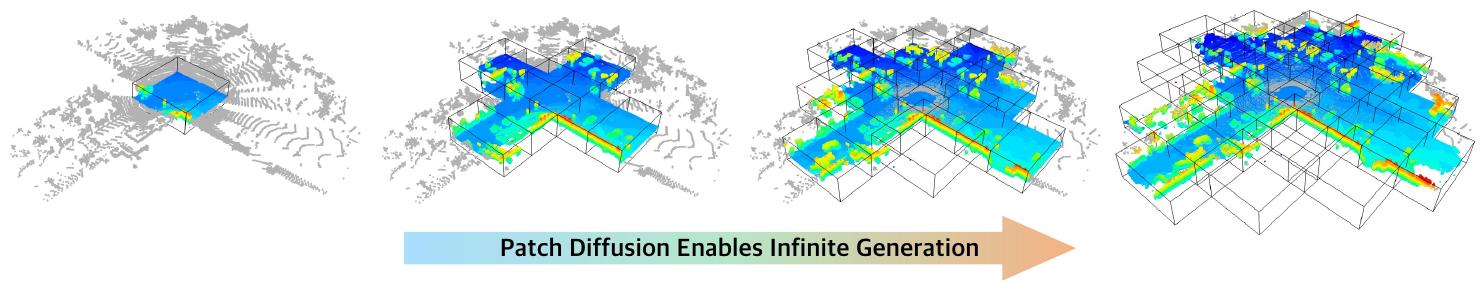}
\captionof{figure}{We introduce PatchScene, a novel diffusion framework based on the divide-and-conquer paradigm, which directly generates point clouds in explicit voxel space. It achieves temporally consistent, high-fidelity, and spatially infinite scene completion through a cycle of patch-based completion, fusion, and diffusion.}
\label{fig:teaser}
\end{center}
}]

\begin{abstract}

We propose PatchScene, a novel diffusion-based framework for large-scale LiDAR scene completion. Unlike existing methods that rely on global latent representations or dense voxel grids, PatchScene adopts a patch-based voxel diffusion paradigm that explicitly generates fine-grained geometry within localized 3D regions. To ensure coherent reconstruction at both spatial and temporal scales, we introduce a confidence-guided spatio-temporal fusion mechanism that integrates overlapping patches and adjacent frames in a unified generative process. Furthermore, we design an Annular-Flow diffusion strategy that leverages the radial density pattern of LiDAR scans to progressively propagate high-fidelity information from near-range to far-range regions, enabling spatially unbounded scene completion. Extensive experiments on the SemanticKITTI benchmark demonstrate that PatchScene achieves state-of-the-art performance across all standard metrics, surpassing previous approaches in both geometric accuracy and temporal consistency. Remarkably, the model trained on 20 m LiDAR ranges generalizes effectively to 50 m scenes without retraining, highlighting its strong scalability and generalization capability for real-world autonomous driving applications. Project page: \href{https://patchscene.github.io/}{https://patchscene.github.io/}

\end{abstract}    
\section{Introduction}
\label{sec:I}

LiDAR is a pivotal sensor in autonomous driving and robotics, providing high-fidelity 3D geometric structures and absolute scale information, which are crucial for robust scene understanding \cite{li2023emergent, meydani2023state}. However, the inherent scanning patterns of LiDAR result in spatially non-uniform point distributions, characterized by density decreasing rapidly with distance \cite{MID, wu2022sparse}. Furthermore, pervasive occlusions inherent to line-of-sight physical measurement prevent the capture of surfaces in occluded regions. These limitations significantly hinder the application of LiDAR point clouds in crucial tasks such as environmental perception, navigation planning, and scene reconstruction for autonomous vehicles \cite{mersch2023building, lidardm}. In contrast, dense point clouds provide richer geometric details and sharper object boundaries, thereby markedly improving the accuracy of downstream algorithms like object detection, tracking, and semantic segmentation \cite{shan2023scp, sanchez2023domain, yi2021complete}. They serve as a fundamental prerequisite for safe navigation and high-precision localization. Nevertheless, sparse-to-dense scene completion remains a formidable challenge in the 3D vision domain, primarily due to the vast scale variations and the irregular, unordered nature of point cloud data in outdoor environments \cite{michele2024saluda, mo2023dit, xiong2023learning}.

A primary challenge in point cloud completion for autonomous driving stems from the large-scale nature of the scenes \cite{liDM, wu2021scene}. This makes achieving efficient generation while maintaining high geometric fidelity exceptionally difficult. To address these challenges, methods such as \cite{lidiff,scorelidar,diffssc} employ techniques like sparse convolution to mitigate the substantial computational overhead. However, the irregular nature of point cloud representations, combined with the difficulty of predicting precise coordinate offsets in continuous 3D space, often results in completed surfaces plagued by artifacts or excessive smoothing, failing to restore sharp geometric structures. XCube~\cite{xcube} attempts to solve the large-scale prediction problem by compressing voxels into a low-dimensional latent space and performing scene completion via latent diffusion, followed by multi-stage refinement. Nonetheless, this multi-stage encoding-decoding process inevitably incurs information loss and error accumulation, leading to degraded geometric detail in the reconstruction \cite{earthcrafter}. A second significant challenge lies in guaranteeing temporal consistency, which is crucial for maintaining continuous target trajectories and smooth geometric surfaces in dynamic environments \cite{tian2023sgtapose}. Yet, prevailing methods are confined to independent, single-frame completion, thereby disregarding the inherent temporal correlations within the sensor data sequence \cite{zyrianov2022learning, yu2021grayscale}.

To overcome these limitations, we propose PatchScene, a novel diffusion framework built upon a divide-and-conquer paradigm. It achieves temporally consistent, high-fidelity, and spatially infinite scene completion through a cycle of patch-based completion, fusion, and diffusion, as illustrated in Fig.~\ref{fig:teaser}. Specifically, we first discretize the full scene and perform diffusion directly within the explicit space of local voxel patches. We introduce an Annular-Flow Patch Diffusion strategy, which guides generation by progressively diffusing outwards in an annular pattern. This effectively propagates dense information from inner, well-observed regions to sparse, distant locations, enabling high-precision completion and extrapolation to infinite spatial domains. Furthermore, we propose a confidence-guided spatio-temporal patch fusion mechanism to enhance inter-frame consistency, realizing coherent and seamless scene completion across temporal sequences. Experimental results demonstrate that our method surpasses previous approaches on the SemanticKITTI dataset \cite{semantickitti}, achieving state-of-the-art (SOTA) performance. The model's ability to train on 20m data and generalize effectively to a 50m range further validates its flexible spatial scalability. Finally, the integration of temporal information allows our method to significantly suppress the inter-frame flickering artifacts common in single-frame approaches, resulting in a substantial quality improvement. 

In summary, our contributions are as follows:
\begin{itemize}
    \item We propose PatchScene, a novel framework that employs diffusion model to explicitly generate fine-grained point clouds in voxel space, flexibly extending to infinite spatial ranges.
    \item We introduce a novel Patch Spatio-temporal Fusion method that ensures holistic continuity and geometric consistency across both spatial and temporal dimensions of the completed point clouds.
    \item PatchScene achieves state-of-the-art results on the SemanticKITTI benchmark, setting a new standard for large-scale  scene level point cloud completion.
\end{itemize}
\section{Related Works}
\label{sec:II}

\begin{figure*}[!t]  
\centering
\includegraphics[width=\textwidth]{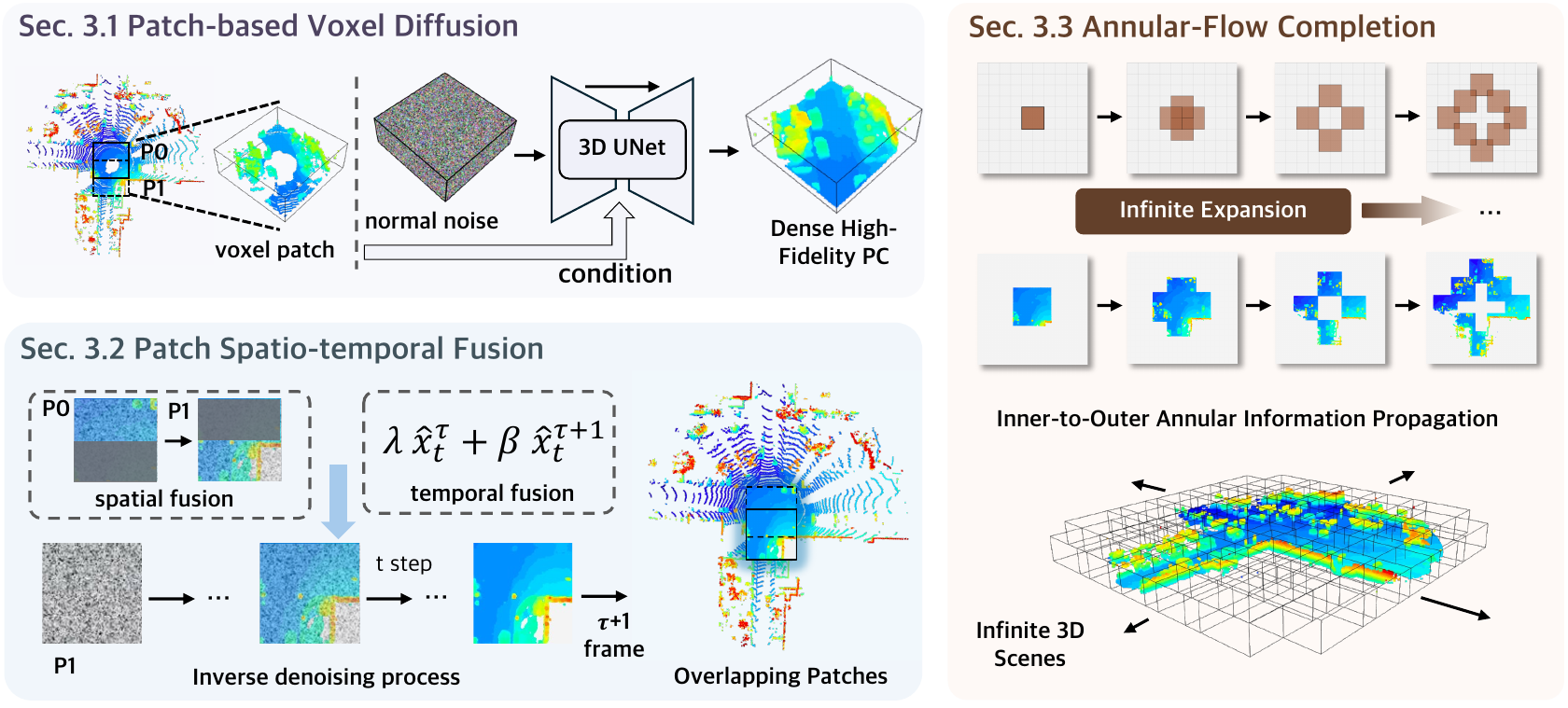}
\caption{Overview of PatchScene. The voxel space is first divided into overlapping local patches, each processed independently through diffusion-based denoising to generate local point clouds. Spatial and temporal fusion then merges these patches into a coherent global point cloud. Finally, an annular outward diffusion strategy extends completion across the entire scene, handling near-dense and far-sparse LiDAR distributions for large-scale, temporally consistent reconstruction.}
\label{fig:PatchScene}
\end{figure*}

\subsection{Discriminative Models for Scene Completion}
\label{sec:2.1}

Early scene completion methods were mostly based on discriminative models,
which learn mappings from partial observations to complete 3D voxel occupancy
or Signed Distance Fields (SDFs) on regular voxel grids, converting completion
into a dense 3D prediction problem. SDF-based methods regress the signed
distance from each voxel to the nearest surface, producing smooth and
structured geometric completions under sparse input conditions
\cite{deepsdf, LODE}. Another mainstream line of work directly predicts the
occupancy state (occupied / free) and semantic labels of each voxel in the 3D
space, leveraging 3D convolutional networks to extract contextual features and
perform end-to-end voxel occupancy prediction, thereby achieving scene-level
semantic completion \cite{LMSCNet, rist2021semantic, roldao20223d, SSCNet}.
Subsequent research introduced Transformer architectures to enhance feature
extraction and fusion capabilities, improving global consistency and
representation quality \cite{voxformer, occ3d}. However, these methods
essentially follow a one-shot prediction mechanism optimized with regression
losses. As a result, they struggle to capture uncertainty and model the
inherent geometric diversity of complex scenes. Related 3D perception tasks have similarly adopted diffusion-based generative modeling to capture plausible hypotheses under partial observations and ambiguities \cite{zhang2023genpose, tian2024robokeygen, zhang2024omni6dpose}.


\subsection{Diffusion Models for 3D Generation}
\label{sec:2.2}
Beyond conditional robotics generation \cite{zeng2024lvdiffusor}, recent studies have extended diffusion models to 3D point clouds. Some works represent LiDAR point clouds as range images, adopting image-generation paradigms to produce diverse and upsampled point clouds \cite{R2DM, liDM, zyrianov2022learning}. However, due to inherent occlusion in range representations, such approaches fail to reconstruct complete 3D scenes. To enable scene completion, subsequent works have performed diffusion directly in 3D space, which can be broadly categorized into three types: point-based, voxel/SDF-based, and latent-based methods.


Point-based methods directly apply diffusion models in the raw point cloud space by predicting offset vectors for noisy points, enabling randomly sampled noise points to move toward missing regions of the scene and thus achieve completion within the point set domain \cite{lidiff, diffssc, lidpm}. Building upon this paradigm, model distillation is introduced in ScoreLiDAR to accelerate and optimize the generative process \cite{scorelidar}. Although operating directly in point space preserves the original geometric structure without voxelization artifacts, the irregular nature of point clouds often leads to unstructured or uneven generation results. Consequently, these methods tend to exhibit structural inconsistency, with holes and oversmoothed fine details in the completed scenes. In voxel/SDF-based approaches, some works primarily applied to single objects or indoor scenes, employing cascaded diffusion models to achieve high-resolution point cloud generation \cite{zhang2025cadgrasp, LasDiffusion, One-2-3-45,One-2-3-45pp}. For outdoor scenes, self-supervised methods like MID recover geometric consistency from sparse LiDAR scans without semantic labels \cite{MID}. However, the memory and computational footprint of explicit 3D grid representations grows cubically, severely limiting the scalability of these methods when applied to large-scale outdoor environments.

For large-scale and high-resolution data generation, latent-based approaches provide a common and effective solution. These methods typically employ a multi-stage VAE framework to encode raw point clouds into a lower-dimensional latent space, where diffusion-based generation is then performed. This strategy enables the synthesis of large-scale and highly dense point clouds \cite{xcube, DiffindScene}. However, such systems are often complex, requiring cascaded VAE training, which is computationally expensive, difficult to optimize, and inherently lossy due to compression.

\section{Method}
\label{sec:III}

To address the challenges of large-scale scenes and temporal multi-view continuous consistency in autonomous driving point cloud completion, we propose PatchScene, which divides the voxel space for local patch generation, followed by spatio-temporal global fusion, and an Annular-Flow Diffusion Completion strategy. The overall architecture is illustrated in the Fig.~\ref{fig:PatchScene}. In Sec.~\ref{sec:3.1}, we introduce the partitioning of the complete voxel space into mutually overlapping regular patches and independently executing a diffusion denoising process to generate local point clouds. Subsequently, in Sec.~\ref{sec:3.2}, we detail our spatial overlap fusion and temporal consistency fusion. These mechanisms simultaneously merge the disparate patches across both temporal and spatial dimensions into a detail-rich, complete, and unified global dense point cloud. Finally, in Sec.~\ref{sec:3.3}, leveraging the physical characteristic of LiDAR scan points of being dense at near ranges and sparse afar, we use an annular, outward progressive diffusion generation method to extend the point cloud completion to infinite spatial domains.

\subsection{Patch-based Voxel Diffusion}
\label{sec:3.1}
Mapping point cloud space directly into a discrete, regular voxel space significantly simplifies the training process and yields more uniform and accurate point cloud completions. However, high-resolution voxelization leads to a cubic growth in computational cost, rendering traditional voxel-based methods infeasible for large-scale scene-level completion. To address this, we partition the entire voxel space using a set of overlapping patches with predefined dimensions (height, width, and depth). We then apply a diffusion model independently to each patch for denoising and completion \cite{karras2022elucidating, luo2021diffusion}. This strategy effectively reduces the task to object-level completion per patch, significantly mitigating the computational and memory burdens. Consequently, we can utilize a high voxel resolution for individual patches, which are subsequently fused spatio-temporally to reconstruct a complete, unified, and high-precision large-scale scene-level point cloud. The diffusion denoising process for our patch-based voxel completion is detailed as follows:

\vspace{0.75em}
\noindent\textbf{Voxel Patching.} We represent a complete, dense occupancy frame as $\mathbf{X}_0 \in \{0, 1\}^{H \times W \times D}$, and its corresponding sparse observation as $\widetilde{\mathbf{X}}$. For the $k$-th patch, we extract a local patch from the global voxel space using predefined dimensions $(h,w,d)$ and a starting index $(i_k,j_k,l_k)$. This patching operation is denoted as:
\begin{equation}
    \mathbf{x}_0^{(k)} = \text{Patch}(\mathbf{X}_0, k), \quad \mathbf{\tilde{x}}^{(k)} = \text{Patch}(\mathbf{\tilde{X}}, k) 
    \label{eq_1}
\end{equation}
where $\mathbf{x}_0^{(k)}, \mathbf{\tilde{x}}^{(k)} \in \{0, 1\}^{h \times w \times d}$. Here, $\text{Patch}(\mathbf{X}, k)$ signifies the $k$-th local patch extracted from the global voxel $\mathbf{X}$. The patches are extracted with a specific stride to ensure sufficient spatial overlap, resulting in a set of $N$ patches that collectively cover the entire scene.

\vspace{0.75em}
\noindent\textbf{Forward Diffusion.} During the forward process in the training phase, noise is independently added to each local patch $\mathbf{x}_0^k$ \cite{DDPM}:
\begin{equation}
    \mathbf{x}_t^k = \sqrt{\bar{\alpha}_t} \mathbf{x}_0^k + \sqrt{1 - \bar{\alpha}_t} \boldsymbol{\epsilon}, \quad t=1, \ldots, T 
    \label{eq_2}
\end{equation}
where $\boldsymbol{\epsilon} \sim \mathcal{N}(\mathbf{0}, \mathbf{I})$ is standard Gaussian noise, and $\bar{\alpha}_t = \prod_{s=1}^t \alpha_s$ are predefined noise schedule coefficients.

\vspace{0.75em}
\noindent\textbf{Inverse Denoising.} In the inverse denoising process, we train a deep neural network $f_{\theta}(\mathbf{x}_t^k, t, \mathbf{p}_k)$ to predict the original data $\mathbf{x}_0$ directly, which enhances the stability and quality of the generated samples. The network accurately recovers the estimated noise-free data $\hat{\mathbf{x}}_0^k$ from the current noisy patch $\mathbf{x}_t^k$ and is conditioned on a learnable position encoding $\mathbf{p}_k$ to ensure the process is sensitive to spatial context.
\begin{equation}
    \hat{\mathbf{x}}_0^k = f_\theta(\mathbf{x}_t^k, t, \mathbf{p}_k)
    \label{eq_3}
\end{equation}
Based on the closed-form solution of the forward diffusion process Eq.~\ref{eq_2}, we substitute the network's prediction $\hat{\mathbf{x}}_0^k$ to derive the corresponding predicted noise $\hat{\mathbf{\epsilon}}$:
\begin{equation}
    \hat{\mathbf{\epsilon}} = \frac{1}{\sqrt{1 - \bar{\alpha}_t}} \left( \mathbf{x}_t^k - \sqrt{\bar{\alpha}_t} \hat{\mathbf{x}}_0^k \right)
    \label{eq_4}
\end{equation}
Subsequently, using this predicted noise $\hat{\mathbf{\epsilon}}$, we can compute the sample for the previous timestep:
\begin{equation}
    \mathbf{x}_{t-1}^k = \frac{1}{\sqrt{\alpha_t}} \left( \mathbf{x}_t^k - \frac{1 - \alpha_t}{\sqrt{1 - \bar{\alpha}_t}} \hat{\mathbf{\epsilon}} \right) + \sigma_t \mathbf{z} 
    \label{eq_5}
\end{equation}
where $\alpha_t = 1 - \beta_t$.

\vspace{0.75em}
\noindent\textbf{Training Objective.} During the training process, for each ground-truth sample, we randomly select a patch $k$ and a timestep $t$. The model $f_\theta$ is then used to predict the corresponding scene patch $\mathbf{\hat{x}}_0^{k}$ from the noisy input $\mathbf{x}_t^k$. We optimize the model by minimizing the Mean Squared Error (MSE) between the prediction $\mathbf{\hat{x}}_0^{k}$ and the ground-truth occupancy voxel $\mathbf{x}_0^{k}$:
\begin{equation}
    \mathcal{L}_{\text{patch}}(\theta) = \mathbb{E}_{k,t}\left[||\mathbf{x}_0^{k} - \mathbf{\hat{x}}_0^{k}||_2^2\right] 
    \label{eq_6}
\end{equation}

\subsection{Patch Spatio-temporal Fusion}
\label{sec:3.2}
\noindent\textbf{Spatial Fusion.} Performing the inverse denoising process independently on all local patches can lead to boundary discontinuities and noticeable artifacts in the overlapping regions, thereby compromising the global consistency of the scene. To resolve this issue, we introduce a stochastic interference spatial fusion mechanism during the inverse diffusion process. This ensures that the denoising outcomes in the overlapping areas of adjacent patches remain smooth and semantically coherent. After obtaining the predicted noise $\hat{\boldsymbol{\epsilon}}^k$ for the current patch at step $t$ according to Eq.~\ref{eq_4}, we project and aggregate the noise predictions from all adjacent patches into a global noise field $\hat{\boldsymbol{\epsilon}}^{\text{global}}$. The fusion operation is exclusively performed within the overlapping regions $\mathcal{O}_{k}$ between patches. For any point $p \in \mathcal{O}_{k}$ within an overlap zone, the fused noise for the next sampling step, $\hat{\boldsymbol{\epsilon}}_{\text{fused}}^k$, is generated through the following probabilistic weighting scheme:
\begin{equation}
    \hat{\boldsymbol{\epsilon}}_{\text{fused}}^k(p) = B(p) \cdot \hat{\boldsymbol{\epsilon}}^{\text{global}}(p) + (1 - B(p)) \cdot \hat{\boldsymbol{\epsilon}}^k(p)
    \label{eq_7}
\end{equation}
where $B \in \{0, 1\}$ is a binary random mask, with each element $B(p)$ drawn from an independent Bernoulli distribution, $B(p) \sim \mathcal{B}(0.5)$. This mechanism mandates that, within the overlapping regions, the noise of the current patch has a $50\%$ probability of adopting the estimate from the global context. Such random coupling effectively integrates local denoising information with the global semantic context, preventing the blurring effects often associated with deterministic fusion methods. This ensures the generation of denoised results with strong spatial coherence in the subsequent inverse step.

\vspace{0.75em}
\noindent\textbf{Temporal Fusion.} While spatial fusion ensures intra-frame coherence, we also aim to fuse information across frames to achieve more natural and geometry-preserving point cloud completions. Consecutive frames are temporally continuous, meaning their observed point clouds are visually coherent and maintain semantic consistency, albeit with slight viewpoint variations. Leveraging the completed dense point cloud from the previous frame can enhance the current frame's understanding of scene semantics and improve the model's implicit grasp of viewpoint information, thereby refining the completion to be detail-rich. During the denoising generation process for frame $\tau$, we cache its prediction at each denoising step. For the subsequent frame $\tau+1$, we first perform ICP \cite{ICP} registration between the sparse observations $\widetilde{\mathbf{X}}^{\tau}$ and $\widetilde{\mathbf{X}}^{\tau+1}$ to obtain the rigid transformation matrix $\mathbf{T}_{\tau \to \tau+1}$, which is used to transform the cached noise into the new coordinate system. When generating the denoised result for frame $\tau+1$ at step $t$, the model relies not only on the observation $\tilde{\mathbf{X}}^{\tau+1}$ but also incorporates guidance from the preceding frame. This temporal fusion is formulated as:
\begin{equation}
    \hat{\mathbf{x}}_{t}^{\tau+1} = \lambda \cdot \hat{\mathbf{x}}_{t}^{\tau} + (1 - \lambda) \cdot \hat{\mathbf{x}}_{t}^{\tau+1}
    \label{eq_8}
\end{equation}
Here, $\lambda \in [0, 1]$ is the cross-frame guidance scale. The value of $\lambda$ is adaptively determined on the Bird's-Eye View (BEV) voxel grid based on local density consistency, dynamically adjusting the weighting between the two consecutive frames.
\begin{equation}
    \lambda(p) = \min\left(\frac{\rho^{\tau+1}(p)}{\rho^{\tau}(p) + \epsilon}, 1.0\right), \quad p \in \mathcal{V}
    \label{eq_9}
\end{equation}
where $\rho(p)$ denotes the local point density at voxel position $p$, and $\mathcal{V}$ represents the set of all voxels in the BEV grid. This design allows the model to inherit more information from the previous frame in geometrically consistent regions while relying more on the current frame's prediction in areas with structural changes or sparsity. This achieves effective temporal fusion across frames in the BEV space.

\subsection{Annular-Flow Diffusion Completion}
\label{sec:3.3}
Given the overlapping nature of our patches, the generation order becomes a critical factor, especially when leveraging completed patches to guide the generation of their neighbors. We observe that 3D LiDAR scenes exhibit a significant physical characteristic: point cloud density is high in proximity to the sensor and becomes progressively sparser with increasing radial distance. Consequently, voxel regions near the sensor are densely populated and relatively complete, whereas distant regions are sparse and often contain large areas of missing data. Since our completion method is conditioned on these observations, the quality of the generated output is inherently higher in the central, denser regions. Therefore, leveraging this physical property, we devise a completion strategy where patches with high-quality completions in the inner regions are used to progressively guide the generation process outwards via spatial fusion, continuing until all patches cover the entire scene. This approach ensures a uniformly dense and globally consistent completion.

Within the scene's voxel space $\mathbf{X}$, we group the patches into a series of concentric annular regions $\{\mathcal{R}_1, \mathcal{R}_2, \ldots, \mathcal{R}_L\}$ based on their distance from the sensor's center. Here, $\mathcal{R}_1$ corresponds to the high-density region near the LiDAR, while $\mathcal{R}_L$ represents the distant, sparse region. Our proposed Annular-Flow process begins with the innermost ring $\mathcal{R}_1$ and proceeds outwards. For each region $\mathcal{R}_\ell$, we perform patch-based conditional diffusion sampling to obtain local completions $\mathbf{\hat{x}}_0^{k}$. Critically, the denoising process for patches in $\mathcal{R}_\ell$ is guided by the already completed results from the adjacent inner ring $\mathcal{R}_{\ell-1}$ at every timestep $t$. This center-outward guided generation allows high-fidelity information to continuously flow from the core to the periphery, enabling the sparse outer regions to leverage the rich semantic context of the inner scene. This center-outward flow endows our method with the inherent capability to extend to unbounded scenes, in principle achieving completion for infinite-scale environments.

\section{Experiment}
\label{sec:IV}

\begin{figure*}[!t]  
\centering
\includegraphics[width=\textwidth]{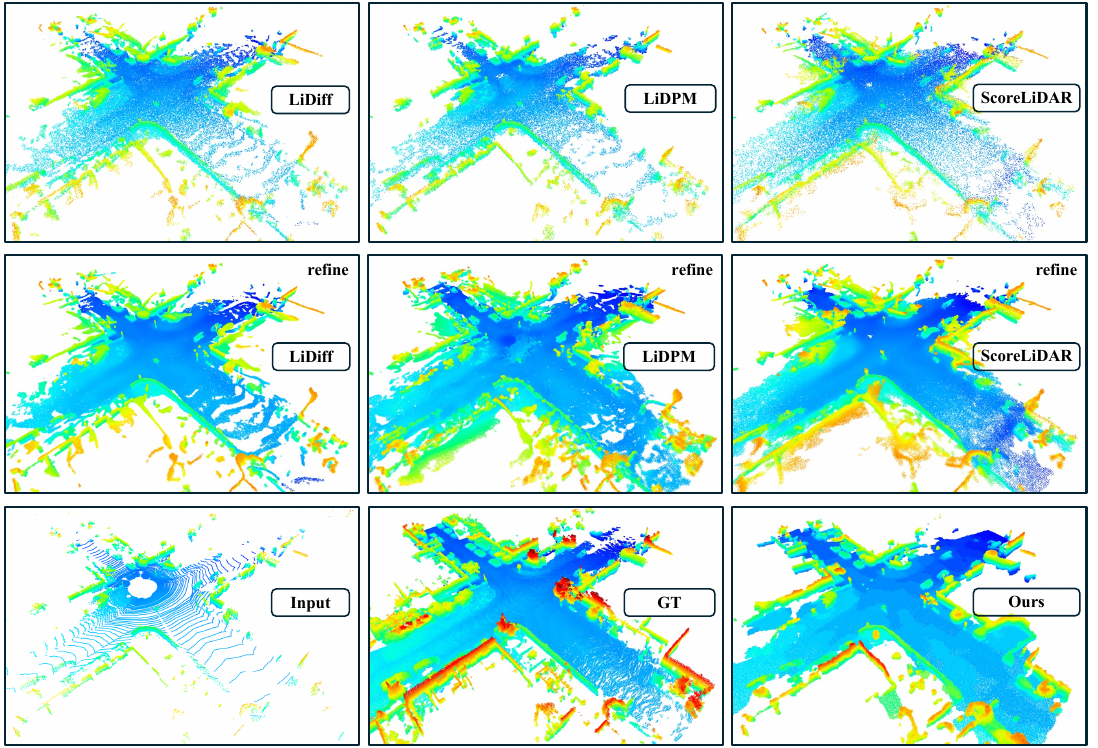}
\caption{Comparison of our method with LiDiff, LiDPM, and ScoreLiDAR on the same scene in SemanticKITTI. All point clouds are height-normalized and color-mapped within the same range. The completion results produced by LiDiff, LiDPM, and ScoreLiDAR remain relatively sparse, and although refinement yields denser predictions, it introduces holes and inconsistent hallucinated points. In contrast, our PatchScene achieves completion results that are closest to the ground truth.}
\label{fig:compare}
\end{figure*}

\subsection{Setup}
\label{sec:4.1}
We conduct training and evaluation on the SemanticKITTI dataset, where the LiDAR scanning range is set to 50\,m, consistent with previous scene-level methods. The voxel resolution is 0.15625\,m. During training, we initialize the learning rate at $4\times10^{-4}$ and employ the AdamW optimizer \cite{adamw} for 100 epochs. For the diffusion process, a cosine noise schedule is applied with $\beta_{10}=0.0001$ and $\beta_{T}=0.02$, using $T=1000$ time steps. When dividing the scene into patches, each patch covers a 20\,m $\times$ 20\,m area. After the denoising generation process, we upsample the completed results by a factor of two, yielding approximately 900,000 points per frame in the final point cloud.

\subsection{Results}
\label{sec:4.2}
Table~\ref{tab_1} provides a comprehensive quantitative comparison between our approach and previous state-of-the-art point cloud completion methods. For fair comparison, all methods are evaluated using the same metric protocols with the original point clouds as ground truth (GT). Our method also does not use any additional temporal information and relies solely on single-frame completion. The evaluation includes Chamfer Distance (CD), Jensen–Shannon Divergence measured in both BEV and full 3D space (JSD-BEV and JSD-3D), and voxel IoU. CD captures the average nearest-neighbor discrepancy between two point sets and reflects local geometric accuracy. JSD-BEV and JSD-3D measure the similarity between global spatial distributions in 2D and 3D, indicating structural plausibility. Voxel IoU evaluates volumetric consistency and large-scale completeness. Based on these complementary metrics, our method demonstrates clear and consistent improvements over all previous approaches, showing its ability to recover fine-grained local geometry while preserving coherent global structure and geometric integrity.

\begin{figure*}[!t]  
\centering
\includegraphics[width=\textwidth]{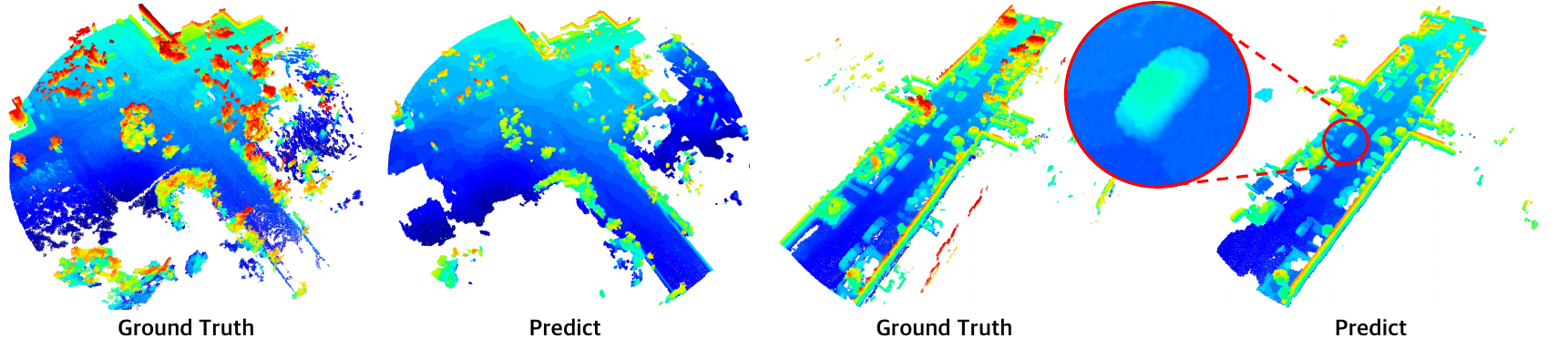}
\caption{We train PatchScene on scenes with a LiDAR sensing range of 20 meters and directly apply it to point cloud completion with an extended range of 50 meters. Whether in open environments or narrow spaces, our completed point clouds consistently maintain high geometric fidelity, accurately preserve object boundaries, and simultaneously ensure strong global scene coherence.}
\label{fig:20m}
\end{figure*}

To further examine the qualitative behavior of different methods, Fig.~\ref{fig:compare} illustrates representative visual comparisons on the SemanticKITTI dataset. LiDiff directly predicts pointwise 3D offsets on raw inputs, which often results in noticeable holes in regions with extremely sparse measurements and produces ambiguous object shapes (e.g., vehicles) due to limited semantic awareness. LiDPM benefits from its global diffusion design, which helps preserve high-level semantic completeness. However, the resulting completions are still coarse near sparse boundaries. ScoreLiDAR generally produces more uniformly distributed completions but frequently introduces hallucinated structures, particularly around object edges, leading to deviations from physically plausible geometry. In contrast, our approach integrates a patch-based voxel diffusion framework with Spatio-Temporal Fusion, enabling more reliable scene-level semantic inference and improved multi-view consistency across temporal sequences. As a result, the generated completions contain substantially fewer scattered noisy points, exhibit stronger structural coherence across both object-level and scene-level regions, and avoid the boundary distortions observed in prior methods. Collectively, these improvements allow our method to achieve clearly superior completion quality in terms of both global scene geometry and local geometric detail.

\begin{table}[!ht]
\centering
\resizebox{\linewidth}{!}{%
\begin{tabular}{lcccccc}
\toprule
\multirow{2}{*}{\textbf{Method}} & \multirow{2}{*}{\textbf{CD$\downarrow$}} & \multirow{2}{*}{\textbf{JSD 3D$\downarrow$}} & \multirow{2}{*}{\textbf{JSD BEV$\downarrow$}} & \multicolumn{3}{c}{\textbf{Voxel IoU$\uparrow$}} \\
\cmidrule(l){5-7}
& & & & 0.5 & 0.2 & 0.1 \\
\midrule
LMSCNet \cite{LMSCNet}      & 0.641 & - & 0.431 & 30.8 & 12.1 & 3.7 \\
LODE \cite{LODE}         & 1.029 & - & 0.451 & 33.8 & 16.4 & 5.0 \\
MID \cite{MID}            & 0.503 & - & 0.470 & 31.6 & 22.7 & 13.1 \\
PVD \cite{PVD}            & 1.256 & - & 0.498 & 15.9 & 4.0  & 0.6 \\
LiDiff \cite{lidiff}         & 0.434 & 0.564 & 0.444 & 31.5 & 16.8 & 4.7  \\
LiDiff(refine)  & 0.376 & 0.573 & 0.416 & 32.4 & 23.0 & 13.4 \\
LiDPM \cite{lidpm}          & 0.446 & 0.532 & 0.440 & 34.1 & 19.5 & 6.3  \\
LiDPM(refine)   & 0.377 & 0.542 & 0.403 & 36.6 & 25.8 & 14.9 \\
ScoreLidar$^\dagger$ \cite{scorelidar}             & 0.412 & 0.589 & 0.425 & 29.7 & 13.0 & 3.6 \\
ScoreLidar(refine)$^\dagger$      & 0.342 & 0.590 & 0.399 & 32.0 & 19.9 & 9.4 \\
\midrule
\textbf{Ours}   & \textbf{0.319} & \textbf{0.444} & \textbf{0.371} & \textbf{45.3} & \textbf{38.2} & \textbf{19.7} \\
\bottomrule
\end{tabular}}
\caption{Comparison of our method with existing approaches on SemanticKITTI. Baselines, metrics, and ground truth are from LiDPM, with results marked ${\dagger}$ independently reproduced and evaluated.}
\label{tab_1}
\end{table}

\subsection{Analysis  of 3D Scene Generation}
\label{sec:4.3}
\noindent\textbf{Temporal Consistency.} To assess temporal continuity, we compute the bidirectional RMSE between adjacent frames. This evaluation captures temporal asymmetries from viewpoint changes or dynamic objects, offering a comprehensive understanding of temporal stability. Table.~\ref{tab_2} presents PatchScene's performance with and without temporal fusion. The results show that temporal fusion significantly reduces both forward and backward RMSE, indicating markedly improved inter-frame coherence and smoother transitions. Crucially, this enhancement in temporal consistency is achieved while largely preserving geometric accuracy. Although there is a marginal increase in JSD BEV, the structural degradation is negligible, with CD and JSD 3D maintain competitive performance. These findings collectively demonstrate that propagating multi-view and temporal information across frames successfully enhances global temporal stability without compromising the overall fidelity of the completed point clouds.

\begin{table}[!ht]
\centering
\resizebox{\linewidth}{!}{%
\begin{tabular}{lccccc}
\toprule
\multirow{2}{*}{\textbf{Method}}& \multirow{2}{*}{\textbf{CD$\downarrow$}} & \multirow{2}{*}{\textbf{JSD 3D$\downarrow$}} & \multirow{2}{*}{\textbf{JSD BEV$\downarrow$}} & \multicolumn{2}{c}{\textbf{RMSE$\downarrow$}} \\
\cmidrule(l){5-6}
& & & & $t_0$ to $t_1$ & $t_1$ to $t_0$ \\
\midrule
w/o temporal       & 0.319 & 0.444 & 0.371 & 0.155 & 0.159\\
temporal fusion    & 0.309 & 0.432 & 0.372 & 0.086 & 0.081\\
\bottomrule
\end{tabular}}
\caption{Analyzing Temporal Consistency with the RMSE Metric}
\label{tab_2}
\end{table}

\vspace{0.75em}
\noindent\textbf{Infinite-Space Generation.} To verify the scalability of our method in infinite spatial extension, we train PatchScene on scenes with a LiDAR range of 20\,m and directly apply it to point cloud completion with a 50\,m range. As illustrated in Fig.~\ref{fig:20m}, we showcase both open and narrow scenes to demonstrate the completion performance of PatchScene under different real-world conditions. In open scenes, the global structure of the completion results produced by PatchScene remains highly consistent with the ground truth. Remarkably, in sparse regions at the right boundary, our method even generates more continuous and complete structures than the ground truth, revealing its strong potential for maintaining global consistency. In narrow scenes, the method effectively completes vehicle point clouds with clear and well-defined shape boundaries, demonstrating that the local diffusion processes, combined with spatial–temporal fusion, preserve both fine-grained details and overall structural integrity. These results indicate that PatchScene can generalize learned scene priors to larger-scale, unseen environments while maintaining reasonable structural accuracy.

\begin{figure}[!t]  
\centering
\includegraphics[width=\linewidth]{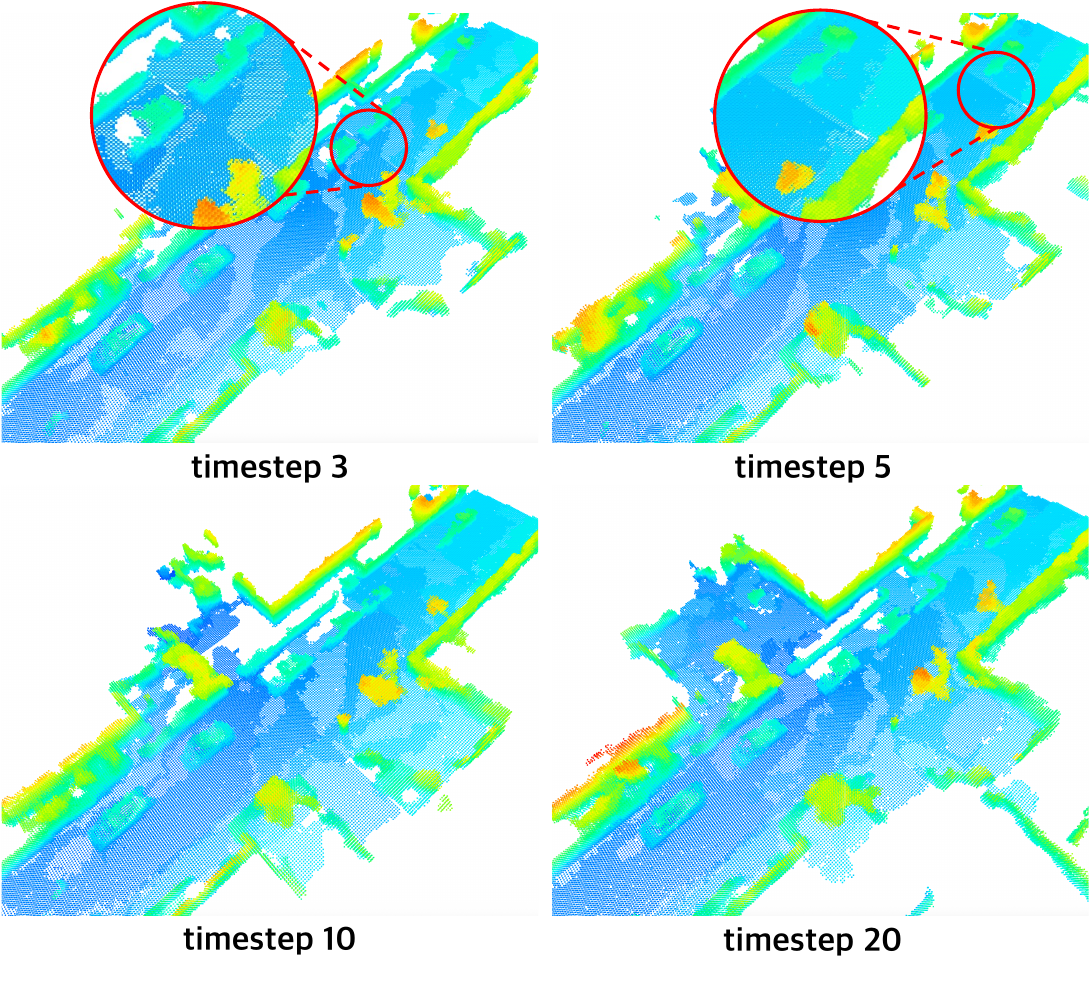}
\caption{The effect of the denoising timestep on completion performance. Less timesteps lead to visible boundaries in the completed scenes, while larger timesteps enable more fusion iterations and yield stronger inter-patch consistency.}
\label{fig:timestep}
\end{figure}

\vspace{0.75em}
\noindent\textbf{Denoising Timestep.} 
We observe that the number of denoising timesteps has a noticeable impact on point cloud completion performance. As shown in Table~\ref{tab_3}, we report the completion metrics under different timestep settings. Interestingly, increasing the number of diffusion steps beyond a moderate value does not necessarily improve performance, since overly large timesteps can degrade quantitative metrics by introducing excessive stochasticity that pushes generated points away from the conditioning inputs. Conversely, using fewer steps can yield slightly higher scores on some metrics, but it limits the effectiveness of patch-wise fusion, leaving visible boundaries between adjacent patches, as illustrated in Fig.~\ref{fig:timestep}. To balance these effects, we adopt 10 timesteps as it provides a favorable trade-off between completion accuracy and inter-patch consistency, achieving the best overall performance across metrics while producing visually coherent reconstructions.

\begin{table}[!ht]
\centering
\resizebox{0.75\linewidth}{!}{%
\begin{tabular}{lcccccc} 
\toprule
\multirow{2}{*}{\textbf{Method}} & \multirow{2}{*}{\textbf{CD$\downarrow$}} & \multirow{2}{*}{\textbf{JSD 3D$\downarrow$}} & \multirow{2}{*}{\textbf{JSD BEV$\downarrow$}} & \multicolumn{3}{c}{\textbf{Voxel IoU$\uparrow$}} \\
\cmidrule(l){5-7}
& & & & 0.5 & 0.2 & 0.1 \\
\midrule
timestep = 5    & 0.310 & 0.437 & 0.371 & 45.7 & 38.9 & 20.3 \\
timestep = 10   & 0.319 & 0.444 & 0.371 & 45.4 & 38.2 & 19.7 \\
timestep = 20   & 0.334 & 0.450 & 0.376 & 44.6 & 37.6 & 19.4 \\
timestep = 30   & 0.349 & 0.453 & 0.378 & 44.3 & 37.3 & 19.2 \\
timestep = 50   & 0.357 & 0.456 & 0.380 & 44.1 & 37.1 & 19.1 \\
\bottomrule
\end{tabular}}
\caption{Analyzing the Impact of Denoising Timesteps on Point Cloud Completion Accuracy}
\label{tab_3}
\end{table}

\subsection{Ablation Study }
\label{sec:4.4}
\noindent\textbf{Spatial Fusion.} To validate the necessity and effectiveness of the spatial fusion design in PatchScene, we conduct a comprehensive ablation study in Table~\ref{tab_4}. Specifically, we evaluate four distinct variants for handling overlapping regions: (i) a naive baseline without spatial fusion; (ii) direct averaging; (iii) weighted averaging prioritizing inner-layer voxels; and (iv) our proposed random coupling strategy. Empirical results reveal that the absence of fusion or the use of deterministic averaging often leads to suboptimal transitions or over-smoothed representations. In contrast, our random sampling strategy, which randomly retains 50\% of overlapping voxels, seamlessly aligns with the underlying statistical properties of the diffusion model. Consequently, it yields the best performance on CD and JSD-BEV metrics while maintaining highly competitive results on JSD-3D, demonstrating its robustness in synthesizing high-fidelity and globally consistent scene completions.

\begin{table}[!ht]
\centering
\resizebox{0.8\linewidth}{!}{%
\begin{tabular}{lcccccc} 
\toprule
\multirow{2}{*}{\textbf{Method}} & \multirow{2}{*}{\textbf{CD$\downarrow$}} & \multirow{2}{*}{\textbf{JSD 3D$\downarrow$}} & \multirow{2}{*}{\textbf{JSD BEV$\downarrow$}} & \multicolumn{3}{c}{\textbf{Voxel IoU$\uparrow$}} \\
\cmidrule(l){5-7}
& & & & 0.5 & 0.2 & 0.1 \\
\midrule
w/o fusion           & 0.348 & 0.451 & 0.383 & 43.9 & 37.4 & 19.7 \\
average addition     & 0.351 & 0.439 & 0.381 & 44.6 & 38.5 & 20.3 \\
weight addition      & 0.345 & 0.438 & 0.379 & 44.9 & 38.6 & 20.3 \\
random coupling      & 0.319 & 0.444 & 0.371 & 45.3 & 38.2 & 19.7 \\
\bottomrule
\end{tabular}}
\caption{Ablation of Spatial Fusion}
\label{tab_4}
\end{table}

\vspace{0.75em}
\noindent\textbf{Generation Direction.} We further investigate the impact of different diffusion directions for patch fusion, as shown in Table~\ref{tab_5}. We compare the completion performance under three diffusion strategies: Annular inward, Annular outward, and Linear diffusion, evaluated using CD, JSD-BEV, and JSD-3D metrics. The Linear Diffusion strategy follows a left-to-right and top-to-bottom propagation manner, which may be intuitive for 2D images but does not align with the radial symmetry characteristic of LiDAR scans, resulting in the poorest performance. Both Annular inward and Annular outward strategies consider the ring-shaped scanning pattern of LiDAR; however, the Annular outward diffusion yields better results, demonstrating the effectiveness of propagating high-fidelity information from the dense central region toward the sparse peripheral areas.

\begin{table}[!ht]
\centering
\resizebox{0.8\linewidth}{!}{%
\begin{tabular}{lcccccc} 
\toprule
\multirow{2}{*}{\textbf{Method}} & \multirow{2}{*}{\textbf{CD$\downarrow$}} & \multirow{2}{*}{\textbf{JSD 3D$\downarrow$}} & \multirow{2}{*}{\textbf{JSD BEV$\downarrow$}} & \multicolumn{3}{c}{\textbf{Voxel IoU$\uparrow$}} \\
\cmidrule(l){5-7}
& & & & 0.5 & 0.2 & 0.1 \\
\midrule
linear diffusion   & 0.451 & 0.528 & 0.399 & 38.2 & 28.7 & 13.0 \\
Annular inward     & 0.391 & 0.461 & 0.389 & 43.3 & 36.7 & 19.0 \\
Annular outward    & 0.319 & 0.444 & 0.371 & 45.3 & 38.2 & 19.7 \\
\bottomrule
\end{tabular}}
\caption{Ablation of Generation Direction}
\label{tab_5}
\end{table}

\section{Conclusion}
\label{sec:V}
In this work, we introduced PatchScene, a diffusion-based framework that addresses the long-standing challenges of large-scale LiDAR scene completion—namely, the trade-off between geometric fidelity, temporal coherence, and computational efficiency. By decomposing the global voxel space into overlapping local patches, PatchScene enables high-resolution completion within a tractable computational budget. The proposed spatio-temporal fusion mechanism ensures globally smooth and temporally consistent reconstructions, while the Annular-Flow diffusion process naturally aligns with the physical scanning characteristics of LiDAR, allowing information to flow outward and support infinite-range scene synthesis.

Through extensive evaluations and ablation studies, PatchScene demonstrates substantial improvements over state-of-the-art methods on the SemanticKITTI dataset, achieving superior Chamfer Distance, JSD, and voxel IoU scores. The framework’s ability to generalize across different spatial scales further validates its robustness and scalability.
{
    \small
    \bibliographystyle{ieeenat_fullname}
    \bibliography{main}
}


\end{document}